%% file: main.tex
\pdfoutput=1
\documentclass[11pt]{article}

\usepackage[final]{acl}

\usepackage{times}
\usepackage{latexsym}
\usepackage{amsmath,amsfonts,amssymb}
\usepackage[T1]{fontenc}
\usepackage[utf8]{inputenc}

\usepackage{microtype}

\usepackage{inconsolata}

\usepackage{graphicx}
\usepackage{hyperref}
\usepackage{url}
\usepackage{booktabs}
\usepackage{longtable}
\usepackage{lineno}
\usepackage{caption}
\usepackage{hyperref}       % hyperlinks
\usepackage{url}            % simple URL typesetting
\usepackage{booktabs}       % professional-quality tables
\usepackage{amsfonts}       % blackboard math symbols
\usepackage{nicefrac}       % compact symbols for 1/2, etc.
\usepackage{microtype}      % microtypography
\usepackage{xcolor}         % colors
\usepackage{wrapfig}
\usepackage{geometry}
\usepackage{caption}
\usepackage{subcaption}
\usepackage{lscape}
\usepackage{float}
\usepackage{longtable}
\usepackage{array}
\usepackage{colortbl}
\usepackage{pdflscape}
\usepackage{tabu}
\usepackage{threeparttable}
\usepackage{threeparttablex}
\usepackage[normalem]{ulem}
\usepackage{manyfoot}%

\usepackage{algorithm}%
\usepackage{algorithmicx}%
\usepackage{algpseudocode}%
\usepackage{listings}%

\usepackage{tabularx}
\usepackage{csquotes}
\usepackage[export]{adjustbox}
\usepackage{wrapfig}
\usepackage{geometry}
\usepackage{caption}
\usepackage{subcaption}
\usepackage{lscape}
\usepackage{float}
\usepackage{longtable}
\usepackage{array}
\usepackage{colortbl}
\usepackage{pdflscape}
\usepackage{tabu}
\usepackage{threeparttable}
\usepackage{threeparttablex}
\usepackage[normalem]{ulem}
\usepackage{makecell}
\usepackage{tabularx}
\usepackage{csquotes}
\usepackage[export]{adjustbox}
\usepackage{makecell}
\usepackage{mathbbol}
\usepackage{graphicx}
\usepackage{tabularx}
\usepackage{booktabs}
\usepackage{multirow}
\usepackage{graphicx}
\usepackage{comment}

\title{Measuring Teaching with LLMs}

\author{Michael Hardy \\
  Stanford University \\
  \texttt{hardym@stanford.edu}}

\begin{document}
\maketitle
\begin{abstract}
Objective and scalable measurement of teaching quality is a persistent challenge in education. While Large Language Models (LLMs) offer potential, general-purpose models have struggled to reliably apply complex, authentic classroom observation instruments. This paper uses custom LLMs built on sentence-level embeddings, an architecture better suited for the long-form, interpretive nature of classroom transcripts than conventional subword tokenization. We systematically evaluate five different sentence embeddings under a data-efficient training regime designed to prevent overfitting. Our results demonstrate that these specialized models can achieve human-level and even super-human performance with expert human ratings above 0.65 and surpassing the average human-human rater correlation. Further, through analysis of annotation context windows, we find that more advanced models—those better aligned with human judgments—attribute a larger share of score variation to lesson-level features rather than isolated utterances, challenging the sufficiency of single-turn annotation paradigms. Finally, to assess external validity, we find that aggregate model scores align with teacher value-added measures, indicating they are capturing features relevant to student learning. However, this trend does not hold at the individual item level, suggesting that while the models learn useful signals, they have not yet achieved full generalization. This work establishes a viable and powerful new methodology for AI-driven instructional measurement, offering a path toward providing scalable, reliable, and valid feedback for educator development.\footnote{\href{https://github.com/hardy-education/measuring_teaching_encoders}{https://github.com/hardy-education/measuring\_teaching\_encoders}}
\end{abstract}

\section{Introduction}
\input{sections/1intro}
\section{Related Work and Motivation}

\input{sections/2relatedwork}

\section{Methods}

\input{sections/3methods}

\section{Data and Experiment}

\input{sections/4data_and_exp}

\section{Discussion}

\input{sections/5discussion}

\section{Conclusion}
Rating classroom teaching quality remains a persistent challenge, with both human evaluators and large language models (LLMs) struggling to effectively utilize authentic classroom observation instruments. While general-purpose GPT models have shown limited promise for this task, we developed custom LLMs based on sentence embeddings to overcome the interpretability and scalability limitations of traditional subword tokenization approaches when processing lengthy classroom transcripts. We systematically evaluated five pretrained sentence embedding models using identical training regimes designed to maximize efficiency and minimize overfitting given the scarcity of authentic classroom data. We assessed their ability to capture pedagogically relevant information using established observation frameworks. Our results demonstrate that three embedding models achieved human-level performance, with correlations exceeding 0.65 for CLASS and surpassing human averages for MQI. More mature models increasingly attribute variation to differences at lesson-level rather than utterance-specific features. This finding challenges prevailing single-turn evaluation paradigms in LLM assessment and development, suggesting that improved models capture broader pedagogical patterns in long-context classroom dynamics. Validity analysis using teacher value-added measures revealed that while models achieving better human alignment also showed stronger alignment with learning outcomes in aggregate, this relationship did not hold at the item level. These results indicate that although models learn pedagogically meaningful features, evidence for generalization remains limited, highlighting important directions for future development of automated teaching quality assessment systems.

\section*{Limitations}

The findings of this study should be considered in light of several limitations related to the data, models, and readiness for practical application. We position this work as a proof of concept, and the following factors must be addressed before these methods can be considered for real-world implementation.

\paragraph{Scope of Data and Generalizability}
The primary limitation of this study is the specificity of the dataset, which consists of transcripts from fourth and fifth-grade mathematics classrooms in the United States. This narrow scope means our models lack proven generalizability to other grade levels, subject areas (e.g., literacy, science), or international school systems. While the underlying methods may hold broader potential, our specific findings are bound to the context of U.S. elementary mathematics education. Expanding the applicability of these models is contingent upon the availability of more varied public data.

\paragraph{Task and Model Specificity}
Our approach is limited by both the evaluation task and the model architecture. We focused on a subset of rating items from the MQI rubric, which may not fully represent the complexity of the universal task of instructional rating. Additionally, the inherent imperfections of observational rubrics, even for human experts, are a constraint on the ground truth data. Furthermore, our encoder models were custom-built for this task. While this specialized design allows a single model to score 25 distinct measures, it is not designed to generalize to new domains or contexts without substantial changes to its architecture or the introduction of new training data.

\paragraph{Considerations for Practical Application and Deployment}
Despite achieving high performance on several metrics, the models in their current state are not ready for high-stakes deployment. Substantial research and validation are necessary to ensure their reliability and to understand potential failure modes. Even when used with a human-in-the-loop, more work is needed to align the models' capabilities with the potential assumptions of end-users. Crucially, this study should not be interpreted as an endorsement for using general-purpose "GPT-style" large language models for similar evaluative tasks. The challenges inherent in this domain require specialized, carefully validated solutions rather than the application of general-purpose technologies.

% \section*{Acknowledgments}

% Bibliography entries for the entire Anthology, followed by custom entries
%\bibliography{anthology,custom}
% Custom bibliography entries only
\bibliography{references}

\appendix

\section{Item Information}\label{apx:iteminfo}

\input{sections/apx_item_information}

\section{Figures}\label{apx:figures}

\input{sections/apx_exp_figures}

\end{document}

%% file: sections/1intro.tex
% \section{Introduction}
Measuring teaching quality is hard \cite{jurenka_towards_2024,kane_gathering_2012,ho_reliability_2013}. Despite their ubiquity as the primary form of teacher development and evaluation, ratings of instructional quality, even when conducted by trained experts, have low reliability, unknown accuracy, and are very expensive to conduct \citep{ho_reliability_2013,kane_national_2015,kane_gathering_2012,glaese_improving_2022,whitehill_automated_2024,whitehurst_evaluating_2014,jurenka_towards_2024, tack_bea_2023,grissom_effective_2013,liu_measuring_2021}. Recent work has sought to reduce the costs of these evaluations using large language models (LLMs) to annotate spoken discourse in classrooms to support such evaluations on actual instruments used with educators, but such models have not yet shown the ability to help with these complex tasks \cite{wang_is_2023,whitehill_automated_2024,xu_promises_2024,hardy_all_2024}. This study builds on these studies, answering calls to do more to evaluate the capacity of LLMs for classroom tasks \citep{casabianca_effect_2013,liu_measuring_2021}.

This study investigates whether pretrained contextual embeddings at the sentence level can meaningfully capture classroom dialogue for automated assessment of instructional quality. We systematically evaluate multitask encoder models trained on fixed sentence-level embeddings to predict expert human ratings of teaching effectiveness across 25 distinct instructional dimensions. We achieve state-of-the-art performance on this task, surpassing existing human benchmark correlations while providing novel insights into the training dynamics of multitask models applied to educational discourse.

Our findings have significant implications for educational assessment and the broader application of NLP methods to specialized domains. We show that shared-weight multitask architectures initially learn general representations of teaching quality that align well with student outcomes, but continued training may lead to overfitting to noisy human annotations on individual constructs rather than the underlying pedagogical constructs of interest. These insights suggest new directions for developing robust AI systems for classroom analysis and highlight fundamental challenges in aligning automated assessments with meaningful educational outcomes.

\subsection{Primary Contributions}

\begin{enumerate}
    \item \textbf{A longitudinal analysis of representation learning for teaching quality}. We provide the first systematic evidence of how different sentence-embedding LLMs develop an understanding of effective instruction throughout their training process.
    \item \textbf{A new benchmark for automated instructional rating}. We demonstrate state-of-the-art performance, outperforming existing models by achieving the highest human-expert correlation across 25 distinct instructional dimensions.
    \item \textbf{A critique of single-turn evaluation}. Through the first analysis of score stability over time, we show that static, single-turn evaluations are insufficient and introduce a more robust, temporally-aware method for assessing LLM-based ratings.
    \item \textbf{A validation framework linking ratings to student achievement}. We introduce the first methodology to directly measure the alignment between an LLM's ratings of teaching and externally-validated measures of student learning gains (teacher value-added).
\end{enumerate}

%% file: sections/2relatedwork.tex
\subsection{Sentence Embeddings}
The robustness of pre-trained language models to out-of-distribution text remains an open question, particularly for specialized domains such as educational settings where child speech patterns and pedagogical discourse structures predominate. To address this challenge, we investigate whether sentence-level embeddings \cite{reimers_sentence-bert_2019} can provide more stable representations of classroom language than traditional subword tokenization approaches, despite the potential loss of fine-grained semantic information. We test the large versions of each of the following pre-trained sentence embeddings: Unsupervised SimCSE and Supervised SimCSE \citep{gao_simcse_2022}, E5 \citep{wang_text_2024}, Multilingual E5 \cite{wang_multilingual_2024}, GTE \citep{li_towards_2023}, and a contrast fine-tuned RoBERTa model released with sentence-transformers \cite{liu_roberta_2019, reimers_sentence-bert_2019}.  This diversity in embedding approaches also allows us to investigate potential biases in model interpretation across different linguistic communities and teaching contexts.

\subsection{Teacher Development and Evaluation}
School leaders working with teachers to improve the quality of instruction typically evaluate the teacher's proficiency in a range of competencies (typically measured during in-class observation and evaluation on a teaching rubric; \cite{aguilar_developing_2013, bambrick-santoyo_get_2016,bambrick-santoyo_leverage_2018}), a process that is often time-consuming and produces ratings (labels) that are unreliable \cite[]{kane_gathering_2012,blazar_validating_2018,kane_have_2013,casabianca_effect_2013}. Without accurate classifications, it is challenging for practitioners to prioritize instructional needs and aligned practices from among the many elements of good teaching \citep{saphier_skillful_2008,darling-hammond_what_2014,hammond_culturally_2015,lemov_teach_2015,lemov_teach_2021,liljedahl_building_2021,darling-hammond_implications_2020,schwartz_abcs_2016} and for researchers to empirically quantify the impact of good teaching practices \citep{pianta_conceptualization_2009,charalambous_13_2019,blazar_challenges_2022,jurenka_towards_2024}. Thus, this work provides a bridge to research seeking to improve teaching quality by providing feedback to teachers on various instructional techniques \citep{samei_domain_2014, donnelly_words_2017, kelly_automatically_2018, demszky_measuring_2021,suresh_talkmoves_2022, jacobs_promoting_2022, alic_computationally_2022,demszky_m-powering_2023, demszky_does_2024, demszky_improving_2023}. 
\paragraph{Automated Evaluation}
Several studies have investigated automated evaluation \cite{whitehill_automated_2024,wang_is_2023,xu_promises_2024,hardy_all_2024}. This study builds on these studies, and replicates the encoder model constructions described by \citeauthor{hardy_all_2025}.\footnote{A more complete description of the model architecture is in \cite{hardy_all_2024}}.

%% file: sections/3methods.tex
% \subsection{Experiment}
For each method, we also display the results at the item-level for better understanding of the learning process.

\subsection{Human Expert Rating Correlation}
% ## Multilevel Partial Spearman's Correlation
To find a generalized measure of correlation across items similar to CLASS and MQI,  we use a multilevel partial Spearman's correlation with inference based on item-level random effects. This accounts for the hierarchical structure while providing a robust, rank-based measure of association that generalizes beyond the specific items sampled. 

$$\rho_{\text{part}} = \text{Corr}(\tilde{R}_{ij}^{(1)}, \tilde{R}_{ij}^{(2)} \mid \mathbf{J}_{ij})$$

\subsection{LLM Rating Stability via Variance Decomposition}
% ## Variance Decomposition Across Model Training Epochs

We employed a generalizability theory framework \cite{brennan_generalizability_2001} to decompose variance in automated LLM scoring across six nested hierarchical levels: sentences (X) within utterances (U) within chapters (C) within lesson stages (S) within lessons (L) within teachers (T), denoted as $X:U:C:S:L:T$. This design enables quantification of context dependency as models evolve during training. The proportion of variance attributable to each level, $h \in \{T,L,S,C,U,X,e\}$, is:

$$\rho_h = \frac{\sigma^2_h}{\sigma^2_T + \sigma^2_L + \sigma^2_S + \sigma^2_C + \sigma^2_U + \sigma^2_X + \sigma^2_e}$$.

\subsection{External Validity via $\tau$-Canonical Correlation Analysis}

% \subsubsection{Canonical Correlation Analysis with Kendall's $\tau$ Kernel}

We employ canonical correlation \cite{hotelling_relations_1936} analysis (CCA) with a Kendall's tau \cite{kendall_new_1938,kendall_treatment_1945} kernel to measure alignment between teacher value-added measures (VAMs) and classroom instructional ratings. In this case, we need a metric that captures the directional alignment only, as differences in scales and ranks may not be meaningful.\footnote{ For example, if two raters give Lesson A the same score of $7$, but give Lesson B different ratings, $3$ and $4$, we would not have evidence to support the notion that LLMs (or even humans) apply an instructional rubric precisely enough for such differences in interval ranks to be practically meaningful when measuring alignment.}  We briefly reconstruct the $\tau$ kernel for creation of scatter matrices here to motivate it as a highly robust \cite{bishara_confidence_2017} measure of alignment between LLM ratings,  $\textbf{x}$, and another metric, $\textbf{y}$ and as having a straightforward alignment interpretation. We translate the following statement into the desired kernel: \textit{"LLM $X$ rates lesson $A$ as better than lesson $B$: $[x_A>x_B]$.  Does the order align with student learning results $Y$ associated with each lesson, $[y_A>y_B]$?"} Thus, for any two lessons,  indexed by $i$ and $j$ and with brackets as indicator functions, we construct this relationship for each:
\begin{align*}\label{eq:paircompairons}
    x_{ij} = [j>i] - [j<i], \quad   y_{ij} = [j>i] - [j<i]
\end{align*}
We construct Gram matrices $\mathbf{K}_X$ and $\mathbf{K}_Y$ where $[\mathbf{K}_X]_{ij} = K_\tau(\mathbf{x}_i, \mathbf{x}_j)$ and analogously for $K_Y$, which are used to solve the eigenvalue problem required for canonical correlation. In our case, all matrices were positive semi-definite and no smoothing was needed. The results are robust to typical transformations for the calculation.\footnote{We also compute the generalized eigenvalue problem using the methods put forward by \cite{yoon_sparse_2020} with the results in the appendix.}

%% file: sections/4data_and_exp.tex
\subsection{Data}
The original classroom lessons used in this study are from the National Center for Teacher Effectiveness (NCTE) Main Study \citep{kane_national_2015},\footnote{\url{https://www.icpsr.umich.edu/web/ICPSR/studies/36095/datadocumentation}} which contains 3 years of data collection and observations of math instruction in approximately 50 schools and three hundred (4th and 5th grade) mathematics classrooms across 4 school districts in the United States. This rich dataset contains \textbf{multiple measures of teaching effectiveness}, including expert ratings of the lessons classrooms, and \textbf{multiple high-quality measures student learning gains}. 

\paragraph{Classroom Lessons and Text}\label{sec:classroomdata}
Human raters watched videos classrooms, and the transcripts\footnote{\url{https://github.com/ddemszky/classroom-transcript-analysis}} of these same videos \citep{demszky_ncte_2022} are used by LLMs for the same task, where the class discourse is equipartitioned across words to align the text with human labels in the absence of timestamps, following \cite{hardy_all_2024}. 

\paragraph{Observation Instruments}\label{sec:mqiratingcriteria}

Our approach encompasses two complementary observation frameworks: a 12-item general teaching practices instrument and a 13-item mathematics-specific teaching assessment \cite[]{bacher-hicks_evaluation_2017, bacher-hicks_experimental_2019}: the CLASS framework \citep{pianta_classroom_2008} for general instructional practice and the content-specific Mathematical Quality of Instruction (MQI) \citep{hill_mathematical_2008}. 

The 63 MQI raters and the 19 external CLASS raters attended biweekly calibration meetings to ensure standardization of scoring procedures. Both frameworks are composed of multiple items that represent distinct instructional dimensions to be evaluated \cite{hill_mathematical_2008, hardy_all_2024,hill_when_2012,kane_gathering_2012}. The MQI and CLASS also represent two types of task for LLMs--detection and summarization, respectively--a distinction that is also clearly illustrated by the distributions of scores coming from the human raters (see Figure \ref{fig:score_variation}). Human rating experts watched videos and provided ratings on all MQI and CLASS items at regular intervals throughout the class, resulting in 779,107 unique numeric ratings provided in 1,762 lessons\footnote{Transcripts are available for 1,600 of the lessons \cite{demszky_ncte_2022}} delivered by 317 teachers to more than 10,000 students in 53 schools.

\paragraph{Value-added measures (VAMs)} are the current gold standard for estimating teacher effects on student achievement gains \cite{kane_gathering_2012, bacher-hicks_evaluation_2017, bacher-hicks_experimental_2019}. VAMs use prior student achievement data and other covariates to measure whether a student's end-of-year performance was above or below the student's expected performance on a standardized exam. The teacher-level VAM is an estimate of the combeined deviations from expected performance for their students, offering a rigorous aggregated estimate of a teacher’s contribution to student learning gains over a school year. As far as we are aware, this is the first study to test LLMs using standardized and value-added measures of student learning. Rare for education datasets, the data of the present study have multiple VAM measurements, which we will use together as a random vector for canonical correlation (and stacked at the year level \cite{bacher-hicks_experimental_2019} for item-level comparisons).

\subsection{Encoder Model Construction}

We develop custom encoder architectures based on sentence-level embeddings to address four key research questions: (1) embedding efficiency in model training, (2) performance relative to human raters, (3) score variation across different temporal contexts, and (4) alignment with rigorous measures of student learning outcomes.

\paragraph{Why Sentence Embeddings? A More Interpretable Architecture.}
Analyzing lengthy classroom discourse poses a challenge for standard LLMs, whose subword tokens are often too granular and computationally intensive for such long-form text. To overcome this, we chose sentence-level embeddings as our foundation. Sentences are natural, interpretable semantic units, allowing our models to efficiently process entire lessons and directly map what a teacher says to established pedagogical frameworks. We evaluated five pretrained sentence-embedding models within a multi-task architecture that learns to score 25 distinct teaching dimensions simultaneously. The model's core weights are shared across all tasks, reflecting the pedagogical theory that effective teaching is a cohesive skill set. Only the final output layers are specialized for each dimension, allowing the model to learn both general and specific features of instructional quality.

\paragraph{Structuring Data for Meaningful Analysis.}
To capture the natural rhythm of a class, we organized transcripts into a three-phase structure (beginning, middle, end) common in elementary math. Each phase was then divided into chapters of a fixed duration (7.5 minutes for MQI, 15 for CLASS). We processed teacher speech into individual sentences, creating uniform inputs and enabling the model to update its assessment with each new utterance. We further augmented our dataset using a sliding-window technique to generate additional training samples from the discourse.

\paragraph{An Efficient and Robust Training Protocol.}
Given the challenge of collecting high-quality observation data, our training protocol prioritized efficiency and generalization. All models were trained for five epochs. We first determined the optimal optimizer by testing AdamW and Adamax; Adamax proved more stable for four of the five models under the strong regularization needed to prevent overfitting. The fifth model (GTE) performed better with AdamW. We retained all models to ensure a comprehensive evaluation. Additional training details follow the protocol in \cite{hardy_all_2024}.

%% file: sections/5discussion.tex
\subsection{Some embeddings are better than others}

\begin{figure}
    \centering
    \includegraphics[width=1\linewidth]{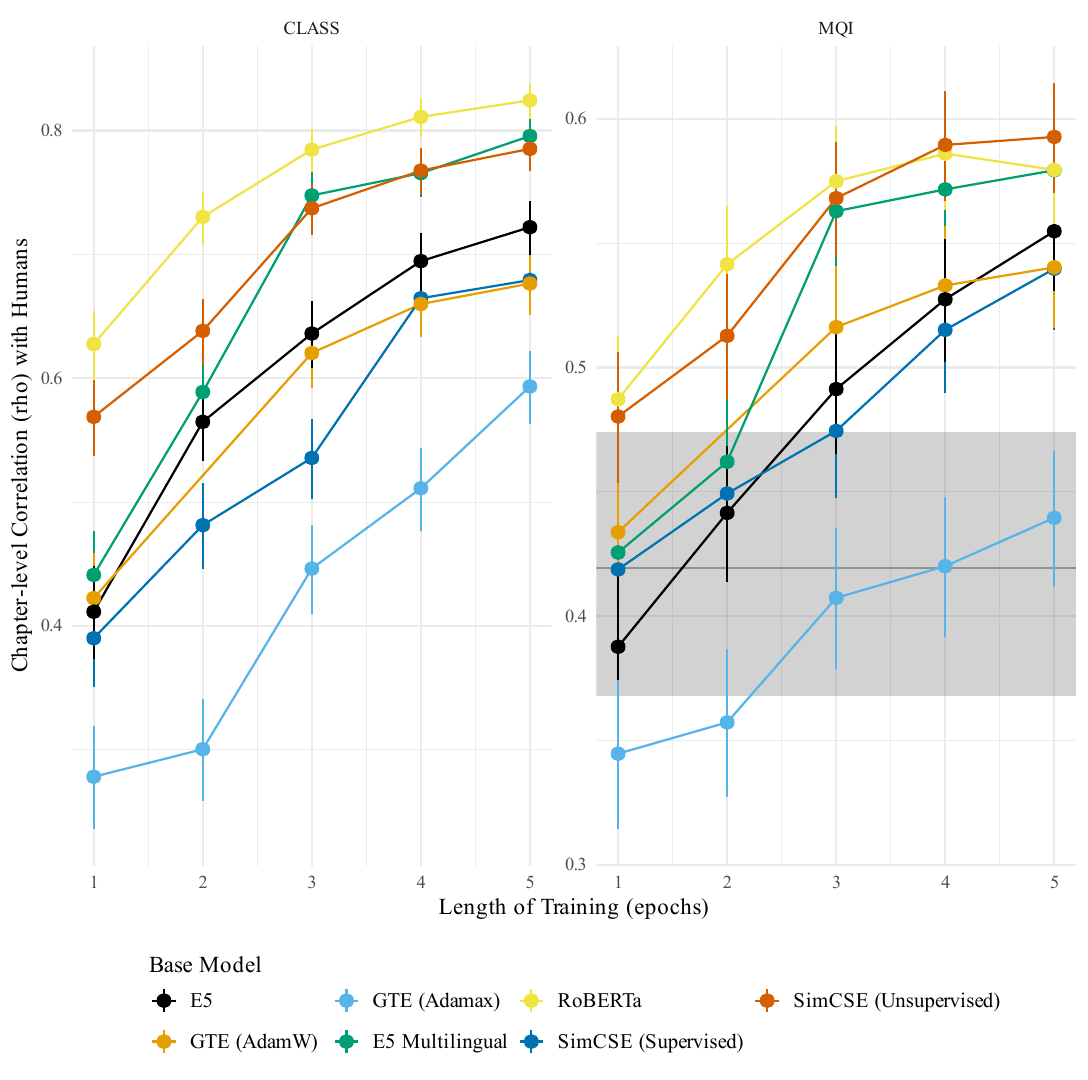}
    \caption{Correlation with human experts across training epochs.  The MQI instrument had at least two human raters per lesson, and the mean and interquartile range of all 63 human MQI raters correlated across the other raters are represented by the gray line and shaded region in the figure. }
    \label{fig:spearman_main}
\end{figure}

Not all contextual embeddings capture the same semantic information. The RoBERTa, Unsupervised SimCSE, and E5 Multilingual models consistently outperformed other embedding models in the present study. When correlations are measured by aggregating to the lesson-level, all models' ratings would be in the top quartile of human raters.  The item level performance at the chapter level can be investigated more deeply in Fig \ref{fig:itemspearmancorrs}.

\paragraph{Implications} Future work should explore fine-tuning SimCSE embeddings or similar self-supervised fine-tuning techniques on related classroom transcripts to investigate the extent to which providing domain-specific contrastive learning could further capture the most relevant semantic information from classroom discourse. Sentence-level embeddings also provide a pathway to interpretability with feature attribution via integrated gradients. Usefulness of feature attribution methods rely on how interpretable each input feature is. In the case of classroom instruction, a sentence spoken is a meaningful unit of discourse, whereas more common methods of creating features for LLMs typically rely on subword tokenization, producing a pixelated semanticity much harder for humans to interpret.

\subsection{Mature models show score stability across longer time windows}\label{sec:score_variation}
\begin{figure}
    \centering
    \includegraphics[width=1\linewidth]{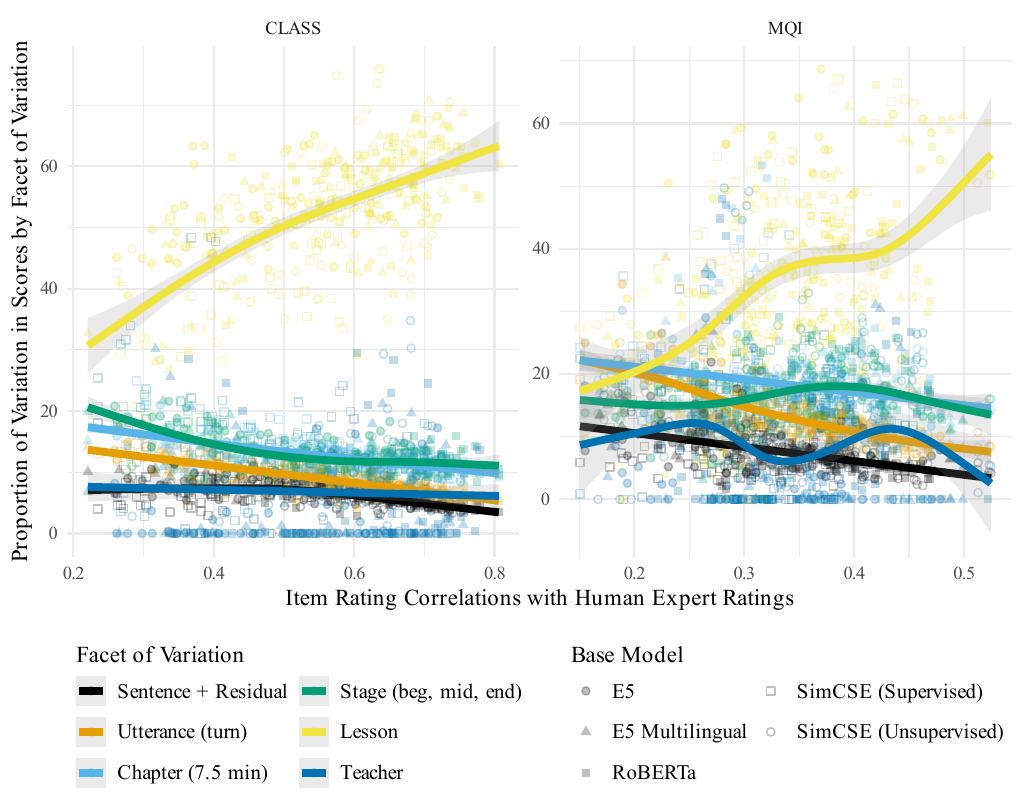}
    \caption{Proportion of variation explained as related to a model's alignment to human expert ratings.}
    \label{fig:score_variation}
\end{figure}

Analysis across training epochs revealed a systematic shift in variance attribution: early-epoch models exhibited substantial utterance-level variation ($\rho_U$), while mature models demonstrated increased lesson-level variation ($\rho_L$) with reduced local context dependency. This finding challenges prevailing single-turn annotation paradigms in LLM evaluation, suggesting that as models improve, they capture broader pedagogical patterns rather than utterance-specific features. Consequently, evaluation frameworks must incorporate extended conversational contexts and hierarchical sampling strategies to accurately assess model performance in educational applications.

We find that the variation in scores for models that are more aligned with human ratings, tends to be less sensitive to smaller changes in time. For the CLASS rubric in particular, more human-aligned models maintain more score stability across an entire lesson, suggesting that the scores are more representative of persistent differences in lessons.

\subsection{External Validity}\label{sec:validity}
\begin{figure}
    \centering
    \includegraphics[width=1\linewidth]{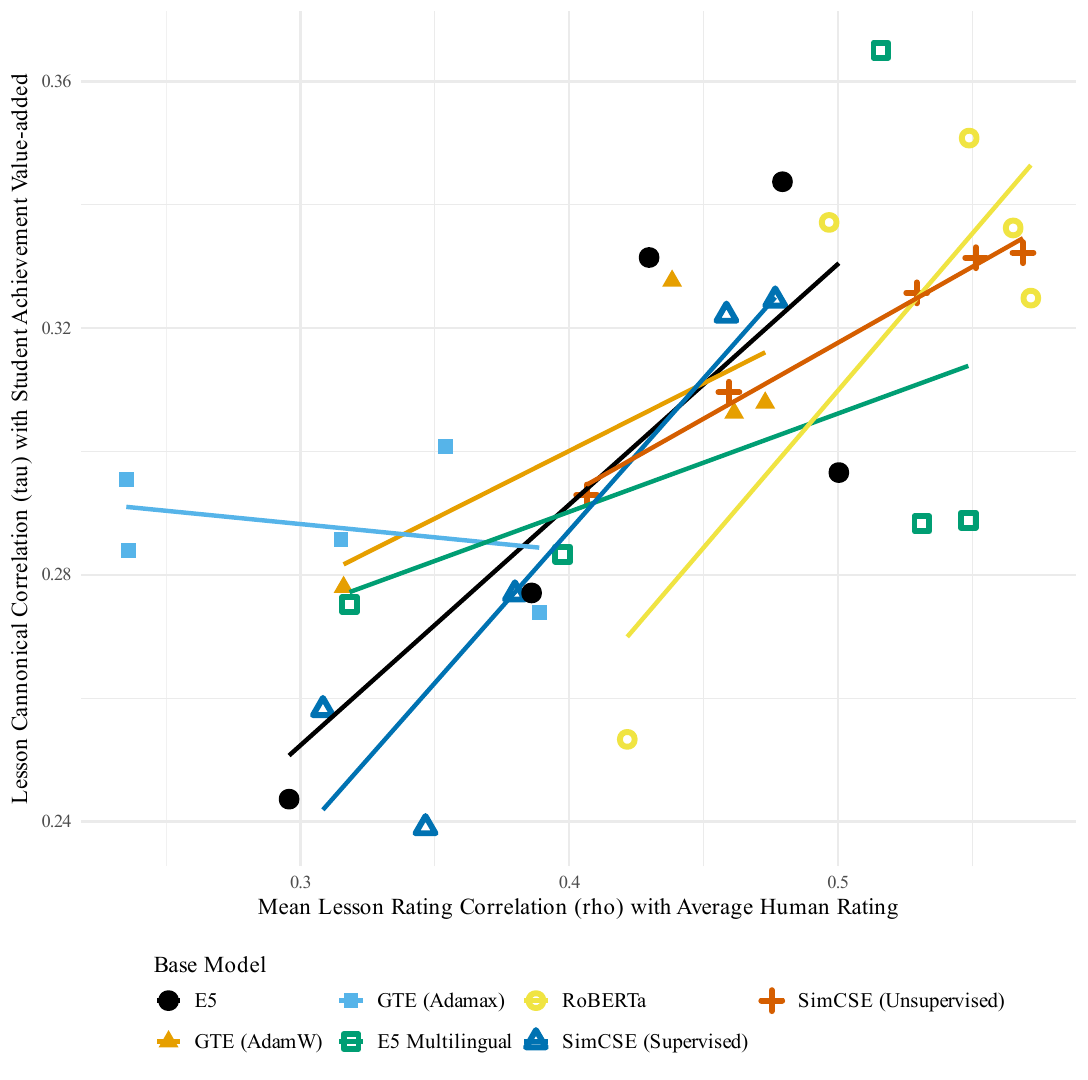}
    \caption{$\tau$-canonical correlation between classroom observation ratings and value-added measures as a function of model alignment to human expert ratings.}
    \label{fig:cancor_rho}
\end{figure}

In order to assess for overfitting  to this particular subset of human raters, we measured the alignment of the chapter-level multi-task scores against the value-added to learning for the students in the class. 
To date, there is no study of which we are aware that connect LLM measures of teaching to the actual value-added in student learning external validity. 

We find that for is increasing alignment between VAM and classroom ratings as models mature and become more aligned with humans.

We demonstrate that while individual task performance generally improves with training, the correlation between model predictions and student achievement outcomes follows a non-monotonic trajectory. This phenomenon reveals important tensions in multitask learning: as models become more specialized at distinguishing between specific instructional skills, their ability to capture the general teaching effectiveness that correlates with student learning gains may paradoxically diminish.

To assess whether our models exhibit overfitting to this particular subset of human raters, we examined the relationship between chapter-level multitask scores and value-added measures (VAM) of student learning outcomes. We employed 
$\tau$-canonical correlation analysis to quantify the strength of association between these two sets of variables while accounting for their multivariate nature.

This analysis addresses a critical gap in the literature: to our knowledge, no prior study has investigated the connection between LLM-derived measures of teaching quality and externally validated VAMs. Such validation is essential for establishing the practical utility of automated classroom assessment tools (Figure~\ref{fig:cancor_rho}).

While the $\tau$-canonical correlation between model predictions and student achievement outcomes generally improves with model sophistication, individual item performance can follow an inverted trajectory performance (Figure \ref{fig:itemwtauVAM}. This phenomenon illuminates an important tension inherent in multitask learning for educational assessment. As models become increasingly specialized at distinguishing between specific instructional skills that human raters prioritize, their capacity to capture the broader dimensions of teaching effectiveness that correlate with actual student learning gains may paradoxically diminish. This suggests that perfect alignment with human expert ratings may not constitute the optimal objective for developing classroom observation tools intended to predict student outcomes.

%% file: sections/apx_item_information.tex
\input{tables/item_descriptions}

%% file: tables/item_descriptions.tex
% TABLE 1
\begin{table*}[h]
% \label{sec:observationinst}
    \centering
    \small
    \setlength\tabcolsep{1.5pt}\renewcommand\defaultaddspace{1.2ex}
    \begin{tabularx}{1\textwidth}{@{} l>{\hsize=0.65\hsize}X>{\hsize=1.30\hsize}X>{\hsize=1.05\hsize\arraybackslash} X @{}}
    \toprule
    \textbf{Abbreviation} & \textbf{Item} & \textbf{Item Description} \\
    \hline
    \hline
    \underline{\textbf{MQI Instrument}} & \\
    ETCA & \textit{Enacted Task Cognitive Activation} & Task cognitive demand, such as drawing connections among different representations, concepts, or solution methods; identifying and explaining patterns. \\
    \textbf{EXPL} &\textit{Teacher Explanations} & Teacher explanations that give meaning to ideas, procedures, steps, or solution methods. \\
    \textbf{LANGIMP}\dag & \textit{Imprecision in Language or Notation} & Imprecision in language or notation, with regard to mathematical symbols and technical or general mathematical language.  \\
    LCP\dag & \textit{Lack of Clarity in Presentation of Mathematical Content} & Lack of clarity in teachers’ launching of tasks or presentation of the content. \\
    LINK & \textit{Linking and Connections} & Linking and connections of mathematical representations, ideas, and procedures. \\
     MAJERR\dag & \textit{Major Mathematical Errors} & Major mathematical errors, such as solving problems incorrectly, defining terms incorrectly, forgetting a key condition in a definition, equating two non-identical mathematical terms. \\
    MGEN & \textit{Developing Mathematical Generalizations} & Developing generalizations based on multiple examples. \\
    MLANG & \textit{Mathematical Language} & Mathematical language is dense and precise and is used fluently and consistently. \\
    MMETH &\textit{Multiple Procedures or Solution Methods} & Multiple procedures or solution methods for a single problem.  \\
    \textbf{REMED} & \textit{Remediation of Student Errors and Difficulties} & Remediation of student errors and difficulties addressed in a substantive manner. \\
    \textbf{SMQR} & \textit{Student Mathematical Questioning and Reasoning} & Student mathematical questioning and reasoning, such as posing mathematically motivated questions, offering mathematical claims or counterclaims. \\
    STEXPL & \textit{Students Provide Explanations} & Student explanations that give meaning to ideas, procedures, steps, or solution methods. \\
    USEPROD & \textit{Responding to Student Mathematical Productions} & Responding to student mathematical productions in instruction, such as appropriately identifying mathematical insight in specific student questions, comments, or work; building instruction on student ideas or methods. \\
    \midrule
    \underline{\textbf{CLASS Instrument}}  & \\
    \textbf{CLPC} & \textit{Classroom Positive Climate} &  \\
    CLNC\dag & \textit{Classroom Negative Climate} &   \\
    CLTS & \textit{Teacher Sensitivity} &  \\
    CLRSP & \textit{Regard for Student Perspective} &   \\
    \textbf{CLBM} & \textit{Behavior Management} &   \\
    CLPRDT & \textit{Productivity} &   \\
    CLILF & \textit{Instructional Learning Formats} &   \\
    CLCU & \textit{Content Understanding} &   \\
    CLAPS & \textit{Applied Problem Solving} &   \\
    CLQF & \textit{Quality of Feedback} &    \\
    \textbf{CLINSTD} & \textit{Instructional Dialogue} &   \\
    CLSTENG & \textit{Student Engagement} &   \\
    \bottomrule
    \end{tabularx}
    \caption{CLASS and MQI item descriptions and corresponding abbreviations. \dag denotes items that are reverse coded due to being negatively worded with respect to the construct of teacher ability. Bolded items are those evaluated by the \textbf{GPT} family of raters and reported by \citeauthor{wang_is_2023}. Each member of the Human and Encoder families of raters evaluated all 25 items.}
    \label{table:items}
\end{table*}

%% file: sections/apx_exp_figures.tex
\subsection{Human Expert Score Distributions}
\begin{figure*}[htbp]
    \centering
    \includegraphics[width=1\linewidth]{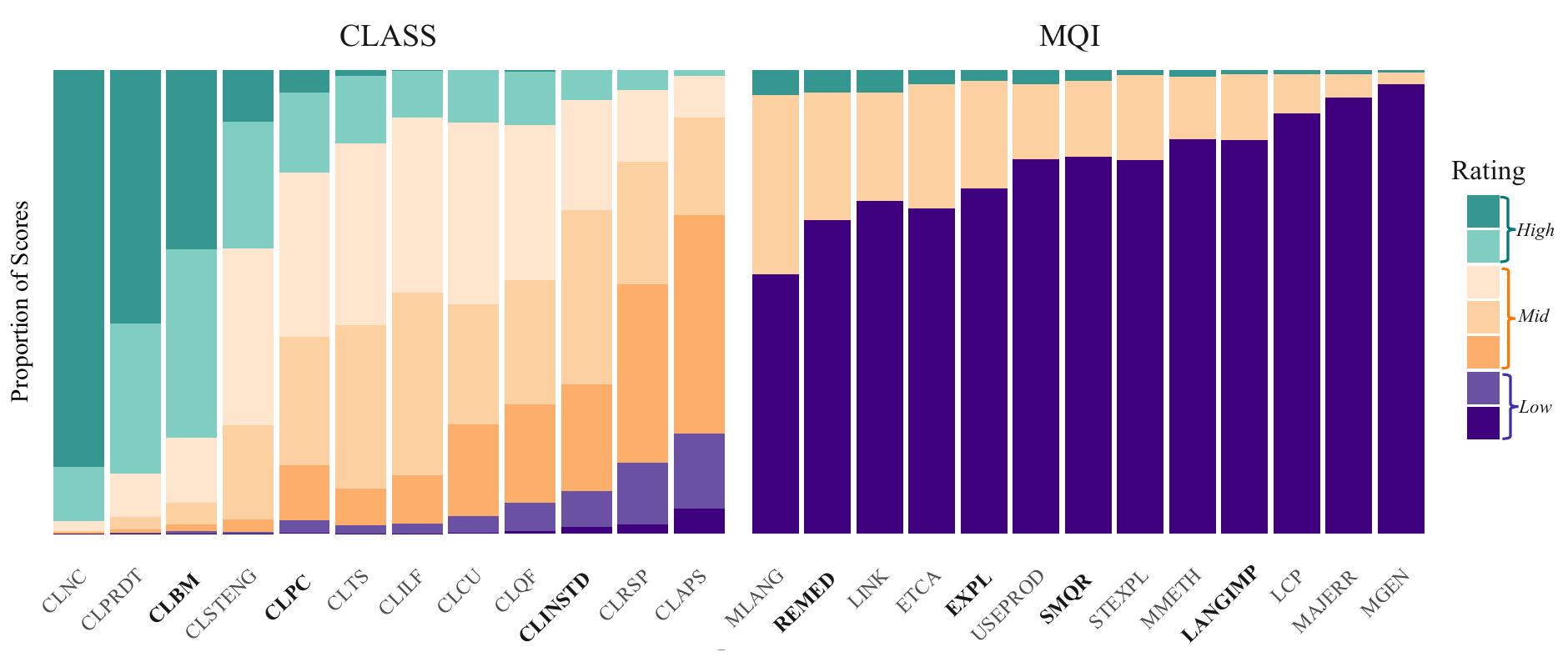}
    
    \caption{\textbf{Human Expert Score Distributions}. These are the score distributions from human experts. The distinct rating patterns highlight the underlying qualitative differences in the constructs being rated. Previous studies have focused on a limited range of items (bolded, \citep{wang_is_2023})}
    \label{fig:llm_vam_raw}
\end{figure*}

\subsection{Test Set Distributions}

\begin{figure*}[htbp]
    \centering
    \includegraphics[width=1\linewidth]{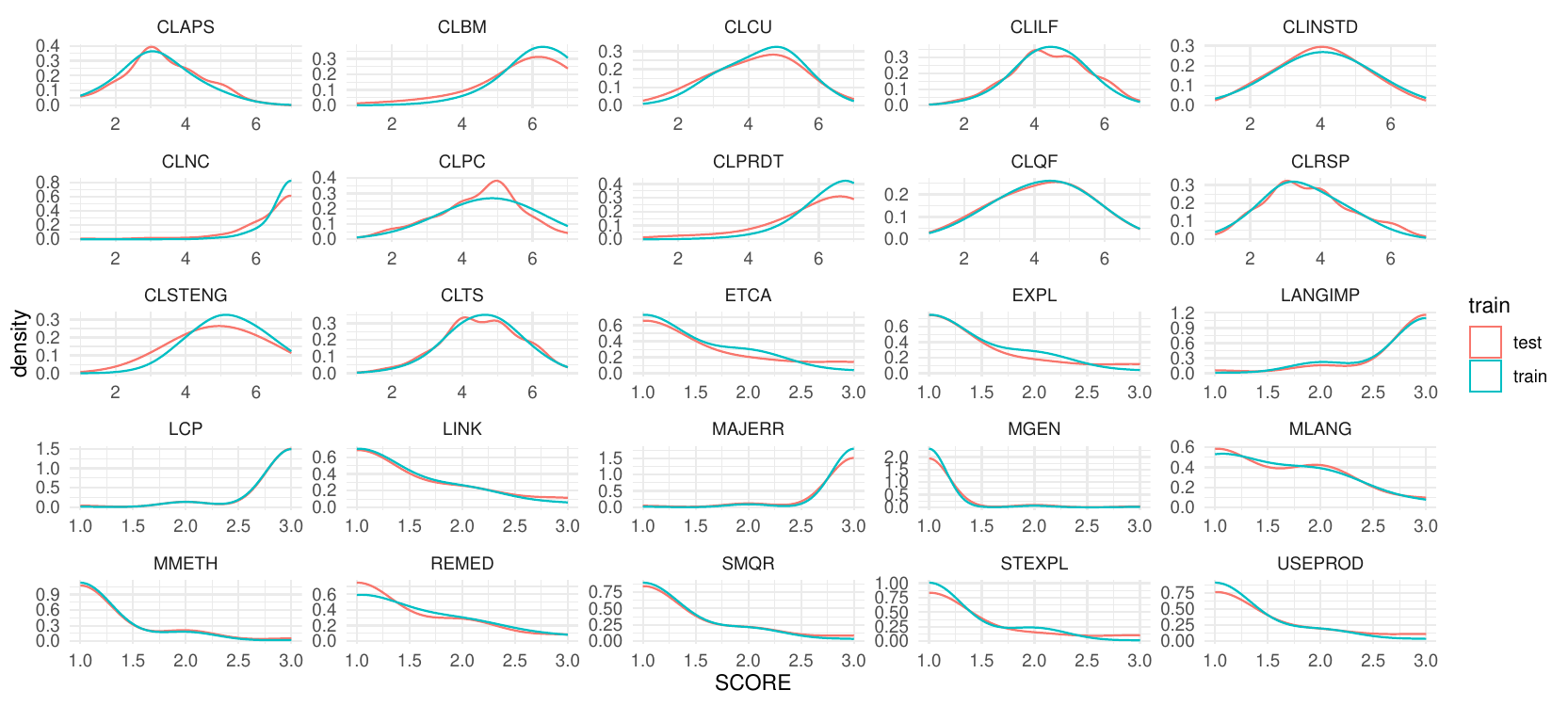}
    
    \caption{\textbf{Held-out Test Set Distributions}. These are comparative score distributions from human experts for the items in the held-out test set and the remaining sample. No differences were statistically significant.}
    \label{fig:llm_vam_raw}
\end{figure*}

\begin{figure*}[htbp]
    \centering
    \includegraphics[width=1\linewidth]{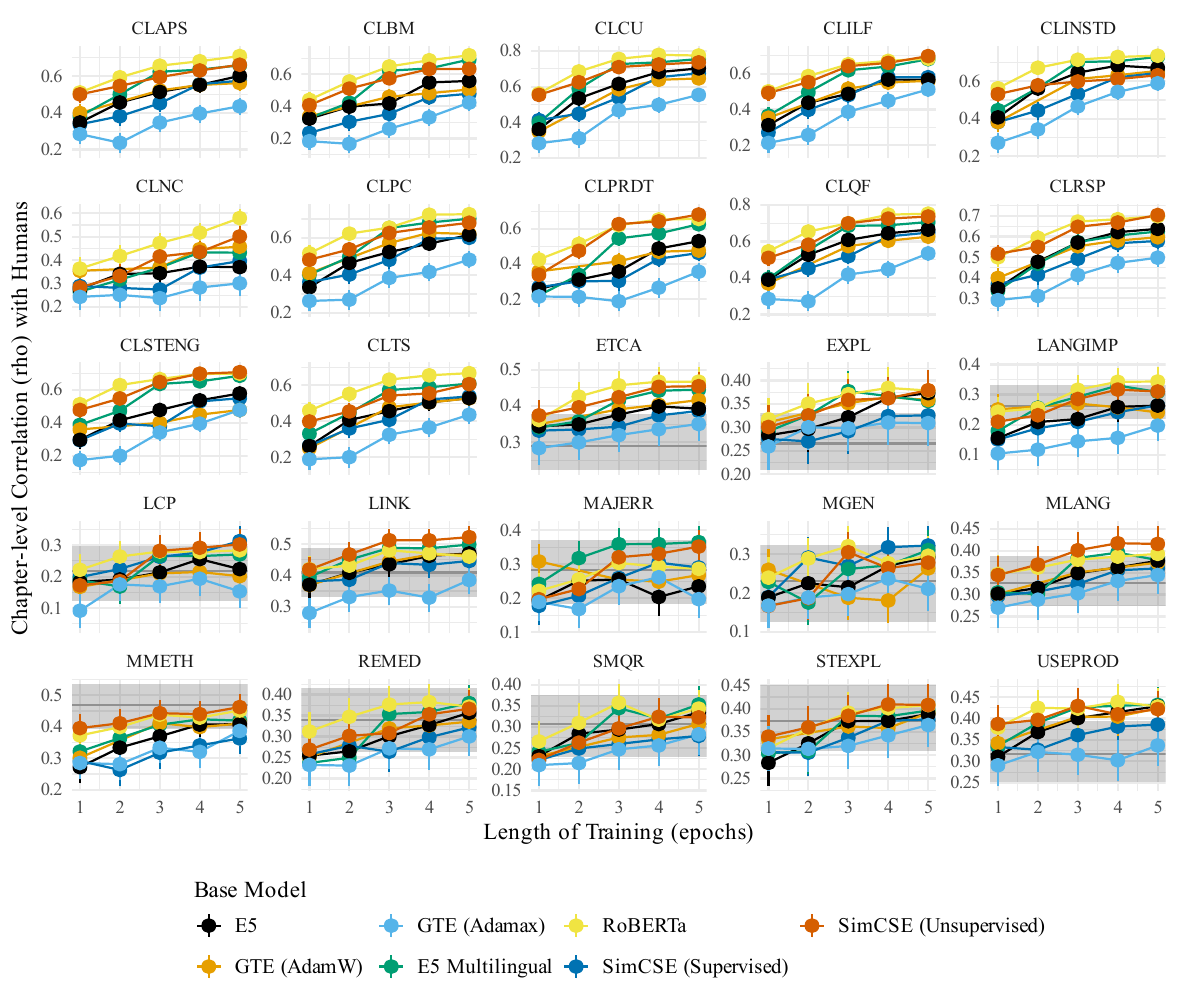}
    \caption{Chapter-level by Item-level Correlation with human experts across training epochs}
    \label{fig:itemspearmancorrs}
\end{figure*}

\begin{figure*}
    \centering
    \includegraphics[width=1\linewidth]{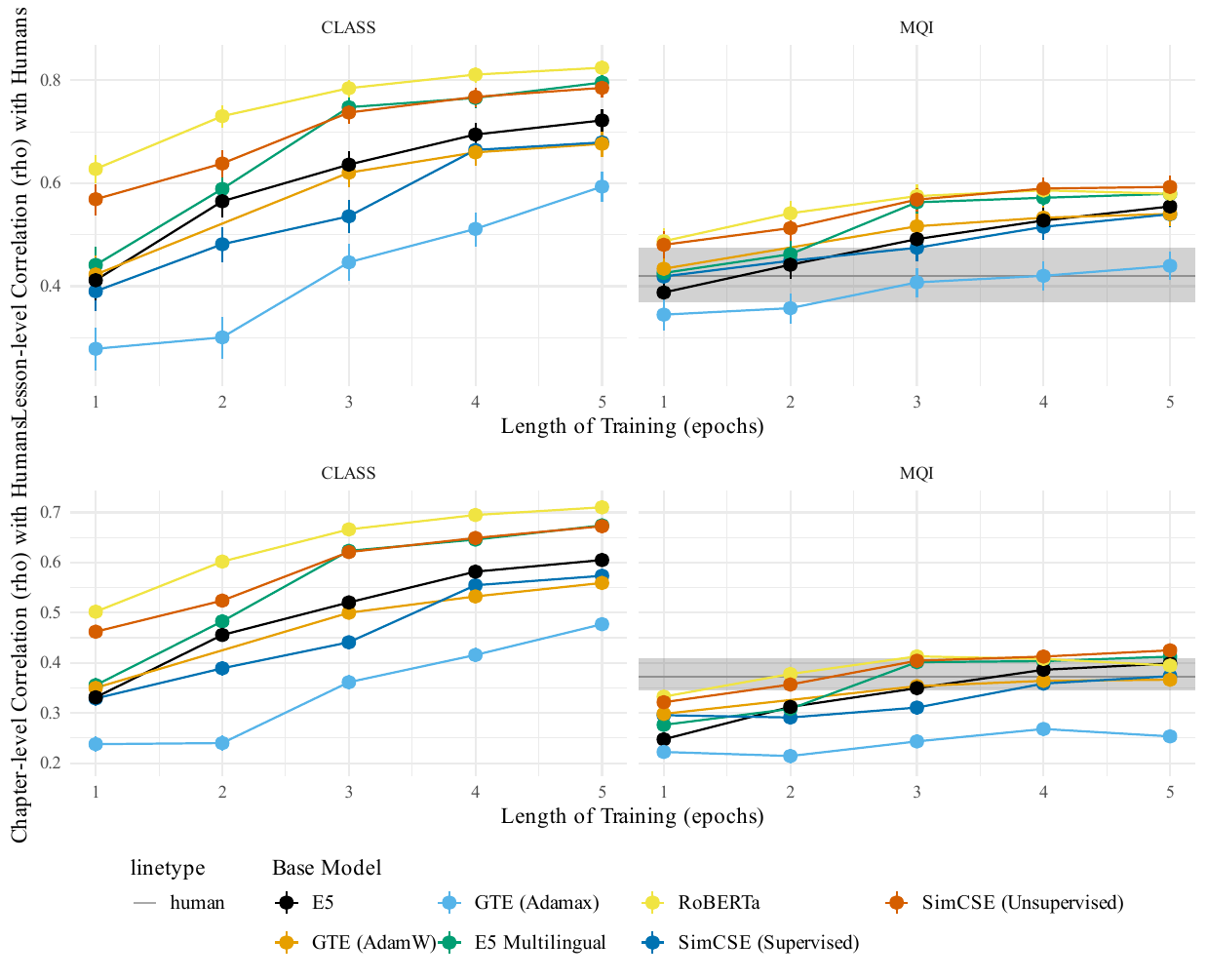}
    \caption{ The MQI instrument had at least two human raters per lesson, and the mean and interquartile range of all 63 human MQI raters correlated across the other raters are represented by the gray line and shaded region in the figure.  When looking at only the specific chapter from the lesson, however, models only perform in the top half of human raters. This discrepancy may be an artifact of the windowing structure used in the training.  The multilevel framework decomposes the correlation into within- and between-cluster components: $\rho_{\text{partial}} = \text{Corr}(\text{rank}(\tilde{R}_{ij}^{(1)}), \text{rank}(\tilde{R}_{ij}^{(2)}) \mid \mathbf{X}_{ij})$ where $\tilde{R}_{ij}^{(k)}$ represents residualized ratings after partialling out covariates $\mathbf{X}_{ij}$ from the multilevel model: $R_{ij}^{(k)} = \boldsymbol{\beta}^{(k)}\mathbf{X}_{ij} + u_j^{(k)} + \epsilon_{ij}^{(k)}$ with random intercepts $u_j^{(k)} \sim N(0, \tau^2)$ for items and residuals $\epsilon_{ij}^{(k)} \sim N(0, \sigma^2)$. The multilevel partial Spearman's correlation accounts for the hierarchical structure while providing a robust, rank-based measure of association that generalizes beyond the specific items sampled, with inference based on the random effects distribution of items.}
    \label{fig:spearman_4panel}
\end{figure*}

\begin{figure*}[htbp]
    \centering
    \includegraphics[width=1\linewidth]{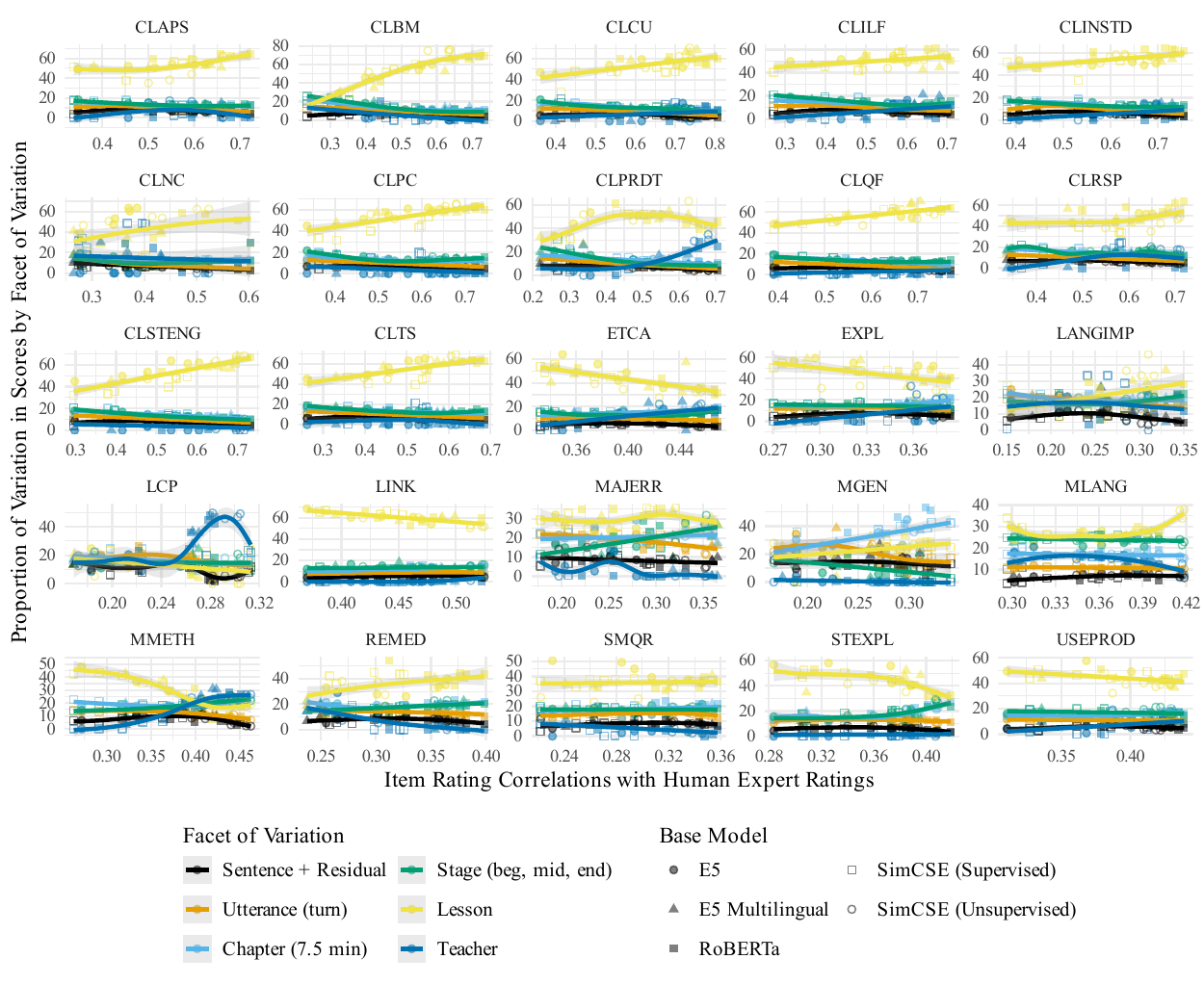}
    \caption{Proportion of variation explained as related to a model's alignment to human expert ratings disaggregated by item using generalizability theory. Random effects were calculated using the \texttt{lme4} package in \texttt{R}: \texttt{SCORE ~ (1|NCTETID) + (1|OBSID) + (1|OBS\_CHAPS) + (1|OBS\_CHAP) + (1|OBS\_CHAP\_idx)}}
    \label{fig:itemvariationfacets}
\end{figure*}

\begin{figure*}[htbp]
    \centering
    \includegraphics[width=1\linewidth]{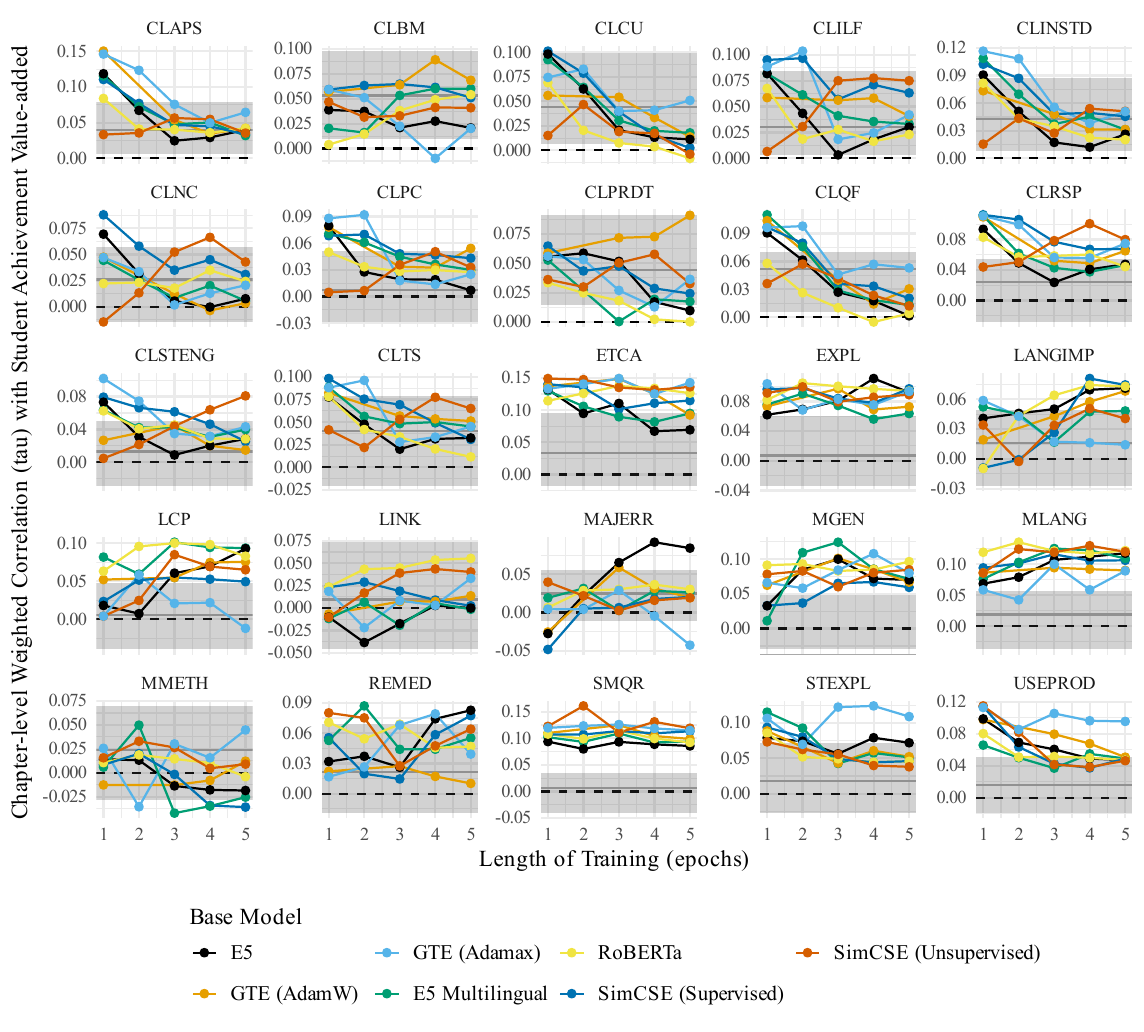}
    \caption{Kendall tau correlations on the stacked VAM outcomes disaggregated at the item level as a function of training epoch. The mean and interquartile range of all human raters evaluated in the same manner are represented by the gray line and shaded region in the figure. Weights are used to account for number of observations per unit.}
    \label{fig:itemwtauVAM}
\end{figure*}

\begin{figure*}[htbp]
    \centering
    \includegraphics[width=1\linewidth]{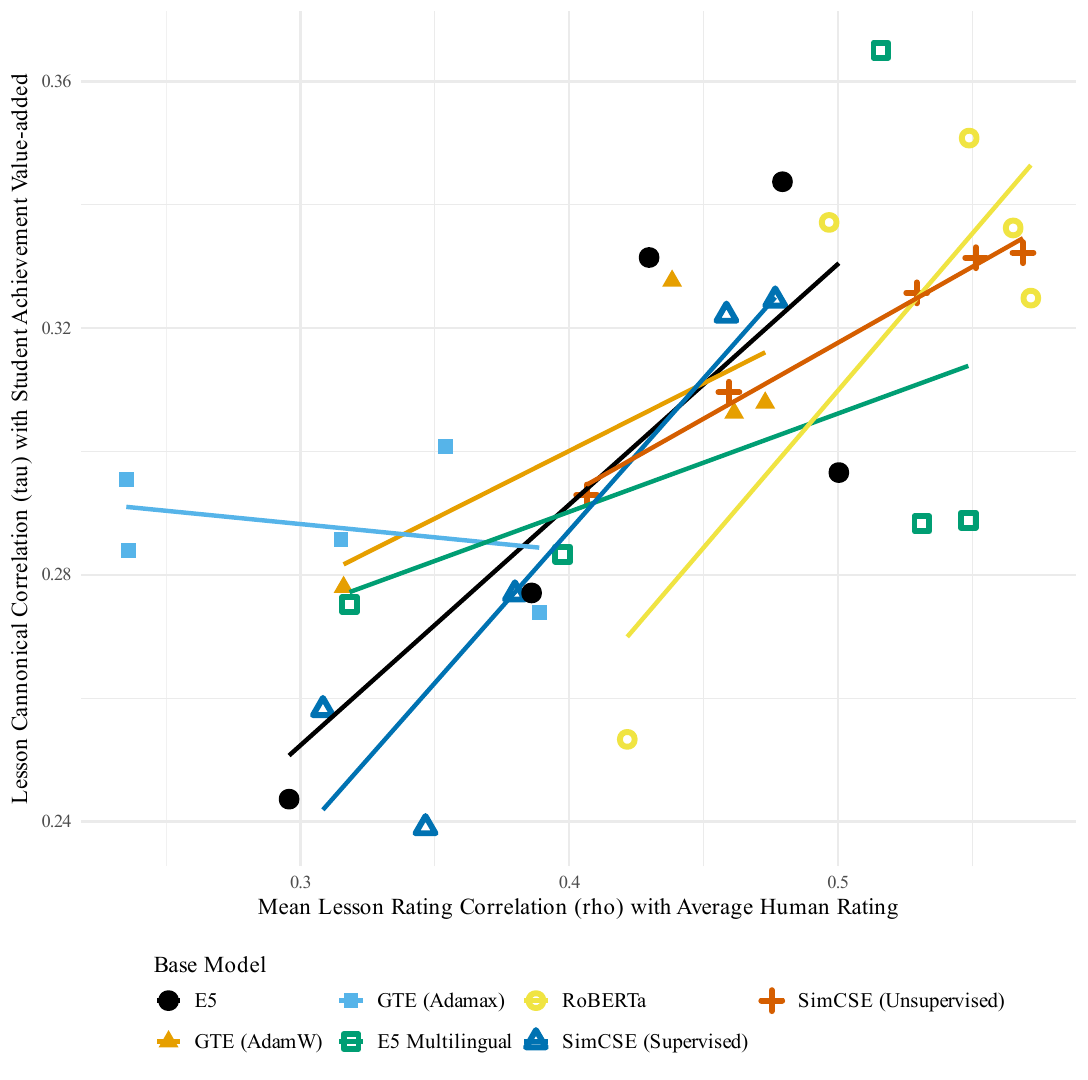}
    \caption{Lesson-level canonical correlations as a function of correlation with human ratings.}
    \label{fig:cancor_lesson}
\end{figure*}

\begin{figure*}[hbp]
    \centering
    \includegraphics[width=0.5\linewidth]{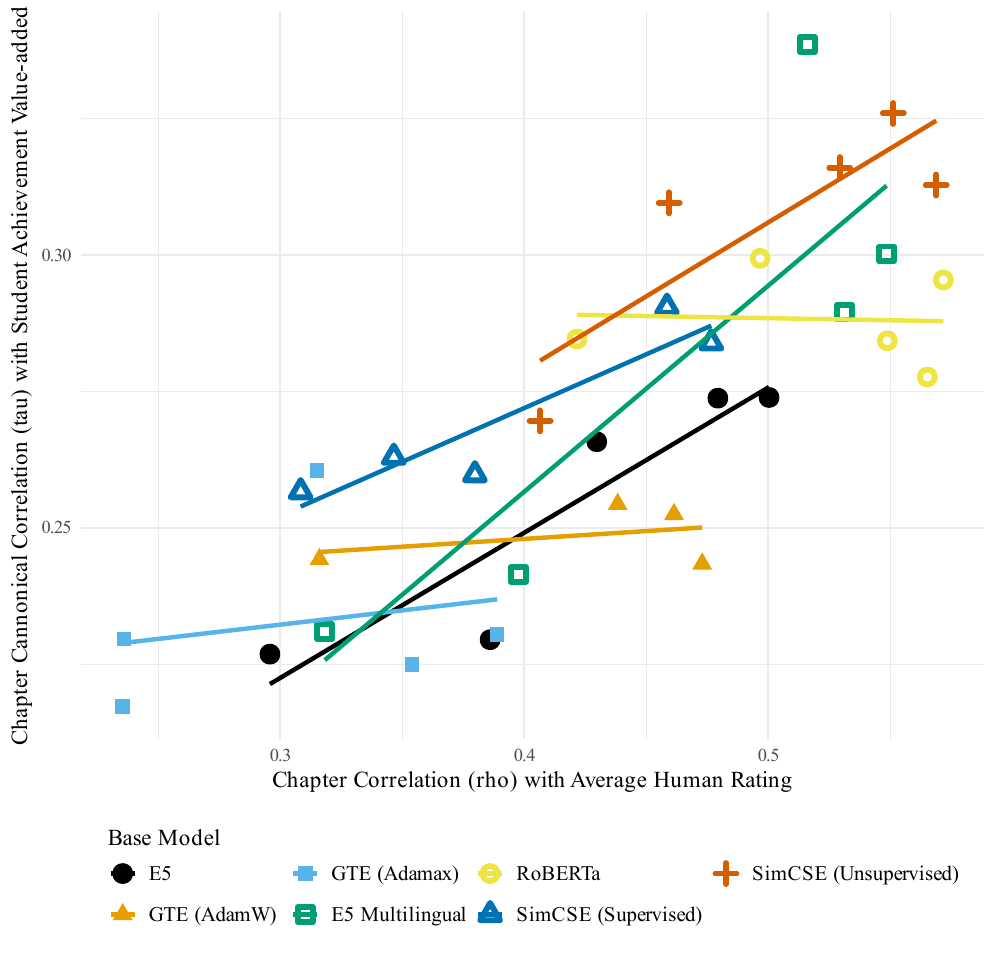}
    \caption{Chapter-level canonical correlations using methods from \citeauthor{yoon_sparse_2020} as a function of correlation with human ratings.}
    \label{fig:mCCA_cancor_chap}
\end{figure*}

\begin{figure*}[hbp]
    \centering
    \includegraphics[width=0.5\linewidth]{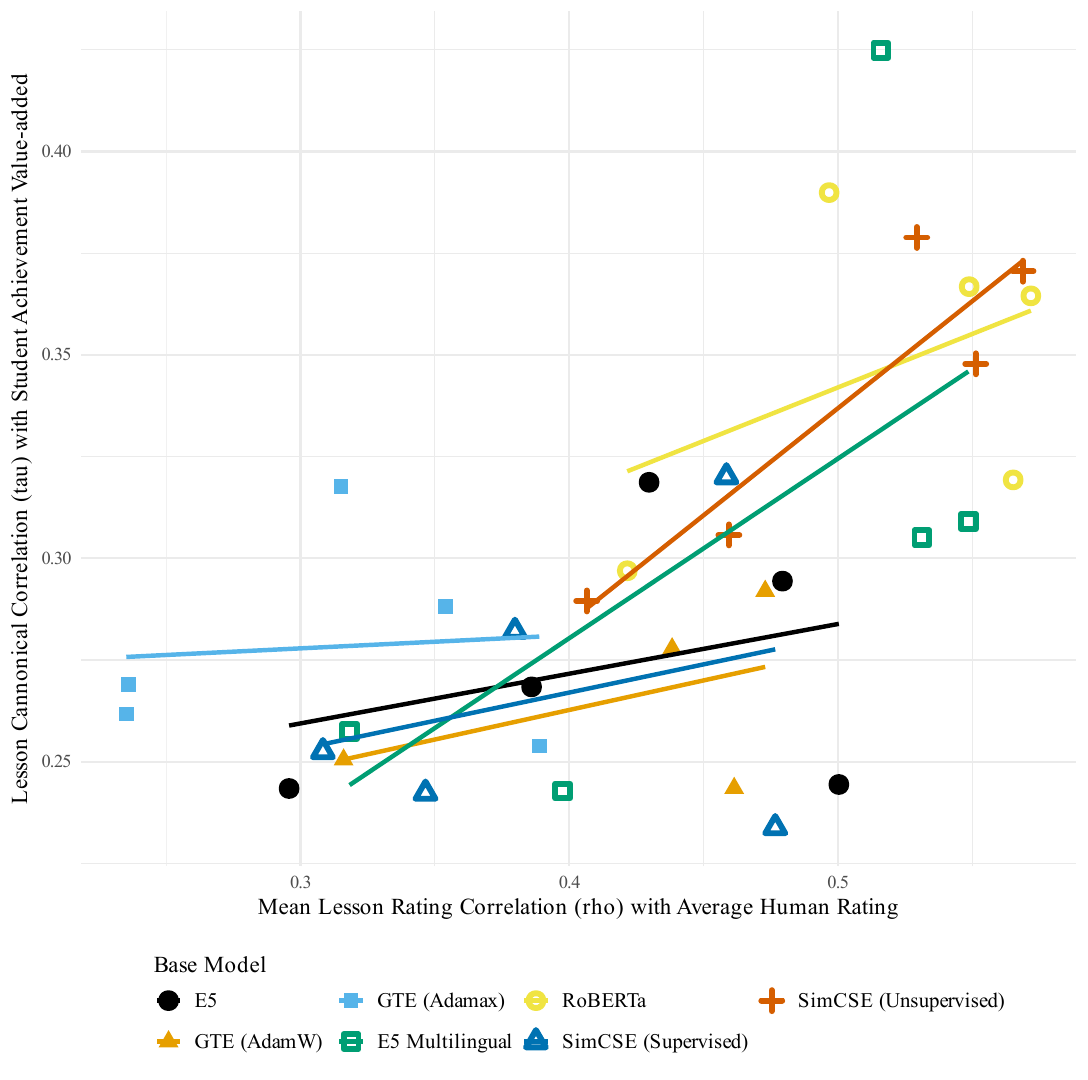}
    \caption{Lesson-level canonical correlations using methods from \citeauthor{yoon_sparse_2020} as a function of correlation with human ratings.}
    \label{fig:mCCA_cancor_lesson}
\end{figure*}